\documentclass[11pt,a4paper]{article}
\usepackage[hyperref]{emnlp2020}
\usepackage{times}
\usepackage{latexsym}
\usepackage{amsmath}
\usepackage{bm}
\usepackage{graphicx}
\usepackage{bbding}

\usepackage{microtype}

\aclfinalcopy 

\newcommand{\argmax}{\mathop{\rm arg~max}\limits}

\title{Multi-Task Learning for Cross-Lingual Abstractive Summarization}

\author{Sho Takase \hspace{1.5em} Naoaki Okazaki \\
  Tokyo Institute of Technology \\
  {\tt \{sho.takase@nlp.c, okazaki@c\}.titech.ac.jp} \\
  }

\date{}

\begin{document}
\maketitle
\begin{abstract}
We present a multi-task learning framework for cross-lingual abstractive summarization to augment training data.
Recent studies constructed pseudo cross-lingual abstractive summarization data to train their neural encoder-decoders.
Meanwhile, we introduce existing genuine data such as translation pairs and monolingual abstractive summarization data into training.
Our proposed method, Transum, attaches a special token to the beginning of the input sentence to indicate the target task.
The special token enables us to incorporate the genuine data into the training data easily.
The experimental results show that Transum achieves better performance than the model trained with only pseudo cross-lingual summarization data.
In addition, we achieve the top ROUGE score on Chinese-English and Arabic-English abstractive summarization.
Moreover, Transum also has a positive effect on machine translation.
Experimental results indicate that Transum improves the performance from the strong baseline, Transformer, in Chinese-English, Arabic-English, and English-Japanese translation datasets.
\end{abstract}

\section{Introduction}
Cross-lingual abstractive summarization is the task to generate a summary of a given document in a different target language~\cite{leuski2003cross-lingual,wan-etal-2010-cross,duan-etal-2019-zero,zhu-etal-2019-ncls}.
This task provides the overview of an article in a foreign language and thus helps readers understand a text written in an unfamiliar language quickly.

Early work on cross-lingual abstractive summarization adopted the pipeline approach: either translation of the given document into the target language followed by summarization of the translated document~\cite{leuski2003cross-lingual} or summarization of the given document followed by translation of the summary into the target language~\cite{orasan-chiorean-2008-evaluation,wan-etal-2010-cross}.
On the other hand, recent studies have applied a neural encoder-decoder model, which is widely used for natural language generation tasks including machine translation~\cite{Sutskever:2014:SSL:2969033.2969173} and monolingual abstractive summarization~\cite{rush-etal-2015-neural}, to generate a summary in the target language from the given document directly~\cite{duan-etal-2019-zero,zhu-etal-2019-ncls}.
Such direct generation approaches prevent the error propagation in pipeline methods.

Training neural encoder-decoder models requires numerous sentence pairs.
In fact, \newcite{rush-etal-2015-neural} provided 3.8M sentence-summary pairs to train their neural encoder-decoder model for English abstractive summarization, and the following studies used the same training data~\cite{zhou-EtAl:2017:Long,kiyono,DBLP:conf/aaai/CaoWLL18}.
However, constructing a large-scale cross-lingual abstractive summarization dataset is much more difficult than collecting monolingual summarization datasets because we require sentence-summary pairs in different languages.
To address this issue, recent studies applied a machine translation model to monolingual sentence-summary pairs~\cite{8370729,duan-etal-2019-zero,zhu-etal-2019-ncls}.
They used the constructed pseudo dataset to train their neural encoder-decoder models.

Meanwhile, the possibility whether existing genuine parallel corpora such as translation pairs and monolingual abstractive summarization datasets can be utilized needs to be explored.
In machine translation, \newcite{dong-etal-2015-multi} indicated that using translation pairs in multiple languages improved the performance of a neural machine translation model.
Similarly, we consider that such existing genuine parallel corpora have a positive influence on the cross-lingual abstractive summarization task since the task is a combination of machine translation and summarization.

In this study, we propose a multi-task learning framework, \textbf{Transum}, which includes machine translation, monolingual abstractive summarization, and cross-lingual abstractive summarization, for neural encoder-decoder models.
The proposed method controls the target task with a special token which is inspired by Google's multilingual neural machine translation system~\cite{johnson-etal-2017-googles}.
For example, we attach the special token \texttt{<Trans>} to the beginning of the source-side input sentence in translation.

The proposed Transum is quite simple because it does not require any additional architecture in contrast to \newcite{zhu-etal-2019-ncls} but effective in cross-lingual abstractive summarization.
Experimental results show that Transum improves the performance of cross-lingual abstractive summarization and outperforms previous methods in Chinese-English and Arabic-English summarization~\cite{duan-etal-2019-zero,ouyang-etal-2019-robust,zhu-etal-2019-ncls,zhu-etal-2020-attend}.
In addition, Transum significantly improves machine translation performance compared to that obtained using only a genuine parallel corpus for machine translation.

Furthermore, we construct a new test set to simulate more realistic situations: cross-lingual summarization with several length constraints.
In a summarization process, it is important to generate a summary of a desired length~\cite{takase-okazaki-2019-positional}.
However, existing test sets for cross-lingual abstractive summarization cannot evaluate whether each model controls output lengths because the test sets do not contain summaries with multiple lengths.
Thus, we translate an existing monolingual abstractive summarization that contains summaries with multiple lengths to construct the new test set.

The contributions of this study are as follows:
\begin{itemize}
    \item We propose a multi-task learning framework, \textbf{Transum}, that uses existing genuine parallel corpora in addition to pseudo cross-lingual abstractive summarization data to train a neural encoder-decoder model.
    \item Transum achieves better scores than previous methods in Chinese-English and Arabic-English summarization. In addition, Transum also improves the performance of machine translation tasks.
    \item We construct a new test set for cross-lingual summarization to simulate more realistic situations. The test set contains summaries with multiple lengths tied to one source document. 
\end{itemize}

\section{Task Definition}\label{sec:definition}
We begin by defining each task addressed in this study before describing our proposed framework.

\subsection{Machine Translation}
The purpose of machine translation is to generate a sentence in a target language from one in a source language while maintaining its content.
Let $V_s$, $V_t$ be the vocabulary of the source and target languages respectively, $\bm{x^s} = (\bm{x_1^s}, ..., \bm{x_M^s})$ be a given sentence in the source language ($\bm{x_i^s} \in \{0, 1\}^{|V_s|}$).
We compute the following $\bm{y^t} = (\bm{y_1^t}, ..., \bm{y_N^t})$, which is the sentence in the target language ($\bm{y_j^t} \in \{0, 1\}^{|V_t|}$):
\begin{align}
  \argmax_{\bm{y^t}} P(\bm{y^t} | \bm{x^s}).
\end{align}

\subsection{Abstractive Summarization}
The purpose of abstractive summarization is to generate a condensed summary of a given sentence.
In addition, we have to preserve a length constraint for the generated summary in real applications~\cite{takase-okazaki-2019-positional}.
Thus, we generate a summary with the desired length $L$.
Formally, we compute the following $\bm{y^s}$ for the given sentence $\bm{x^s}$ and desired length $L$:
\begin{align}
  \argmax_{\bm{y^s}} P(\bm{y^s} | \bm{x^s}, L),
\end{align}
where the length of $\bm{y^s}$ is equal to $L$.

\subsection{Cross-Lingual Abstractive Summarization}
In cross-lingual abstractive summarization, we generate a summary in the target language with the desired length $L$ for a given sentence.
Thus, we compute the following $\bm{y^t}$:
\begin{align}
  \argmax_{\bm{y^t}} P(\bm{y^t} | \bm{x^s}, L),
\end{align}
where the length of $\bm{y^t}$ is equal to $L$; this is the same as in (monolingual) abstractive summarization.

\section{Proposed Method: Transum}\label{sec:transum}
\begin{figure}[!t]
  \centering
  \includegraphics[width=7.5cm]{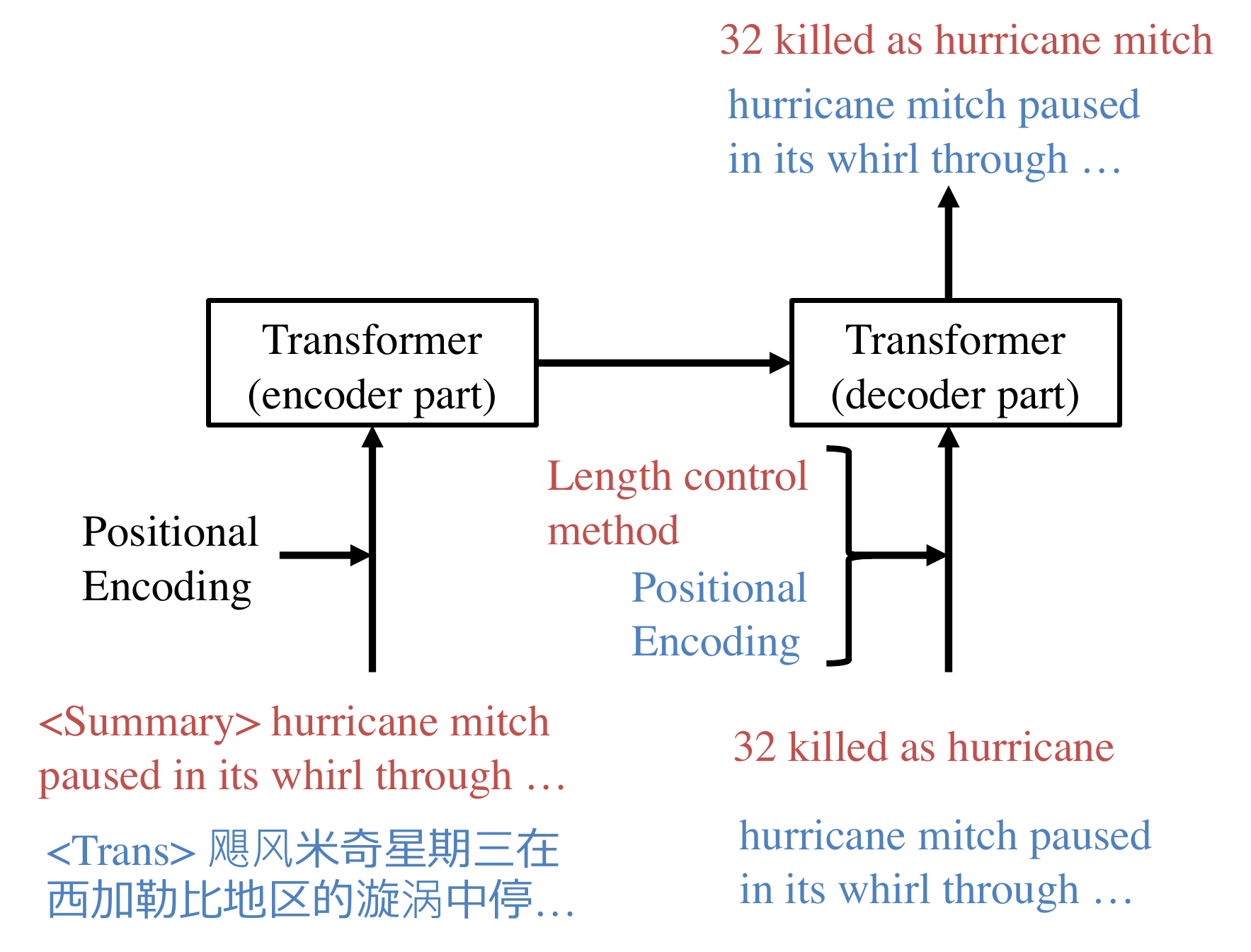}
   \caption{The overview of our proposed Transum. Transum requires attaching the special token \texttt{<Trans>} or \texttt{<Summary>} to the beginning of the input sentence and outputs the sequence based on the attached token. Thus, we can use both translation pairs and monolingual sentence-summary pairs in addition to the cross-lingual abstractive summarization data to train the neural encoder-decoder.}
   \label{fig:overview}
\end{figure}

Figure \ref{fig:overview} shows an overview of the proposed multi-task learning framework: \textbf{Transum} which uses both of translation pairs and monolingual sentence-summary pairs for cross-lingual abstractive summarization.
We attach the special token \texttt{<Trans>} or \texttt{<Summary>} to the beginning of the input sentence as shown in this figure.
We expect to control the output of the decoder with the special token in a similar manner to that used in multilingual translation, where a special token indicates the target language~\cite{johnson-etal-2017-googles}.
Our used special token depends on only the target side.
In other words, we use \texttt{<Summary>} for both of monolingual summarization and cross-lingual summarization.
This configuration enables Transum to process cross-lingual abstractive summarization even if we train it on only translation pairs and monolingual summarization data (i.e., zero-shot generation).
For example, if we replace \texttt{<Trans>} with \texttt{<Summary>} in the input Chinese sentence as shown in Figure \ref{fig:overview}, the decoder should generate a summary in English.

As described in Section \ref{sec:definition}, we have to preserve the length constraint in decoding abstractive summarization.
Thus, we introduce a method to control output sequence length into our encoder-decoder.
In this study, we adopt the length-ratio positional encoding (LRPE)~\cite{takase-okazaki-2019-positional}, which is a variant of the sinusoidal positional encoding~\cite{NIPS2017_7181}, to generate an output with the desired length $L$ in Transformer.
Let $pos$ be the position of an input token and $d$ be the embedding size.
Then, the $i$-th dimension of LRPE, i.e., $LRPE_{(pos, L, i)}$ is as follows:
\begin{align}
  LRPE_{(pos, L, 2i)} &= {\rm sin}\bigg(\frac{pos}{L^{\frac{2i}{d}}}\bigg), \label{eq:ctrl_ratio_sin}\\
  LRPE_{(pos, L, 2i+1)} &= {\rm cos}\bigg(\frac{pos}{L^{\frac{2i}{d}}}\bigg). \label{eq:ctrl_ratio_cos}
\end{align}
Following \newcite{takase-okazaki-2019-positional}, we use LRPE for the decoder side during summarization.
Thus, as shown in Figure \ref{fig:overview}, we adopt LRPE in the case that \texttt{<Summary>} is attached to the input sentence; otherwise, we use the conventional sinusoidal positional encoding.

\section{Pseudo Data Construction}

\begin{figure}[!t]
  \centering
  \includegraphics[width=7.5cm]{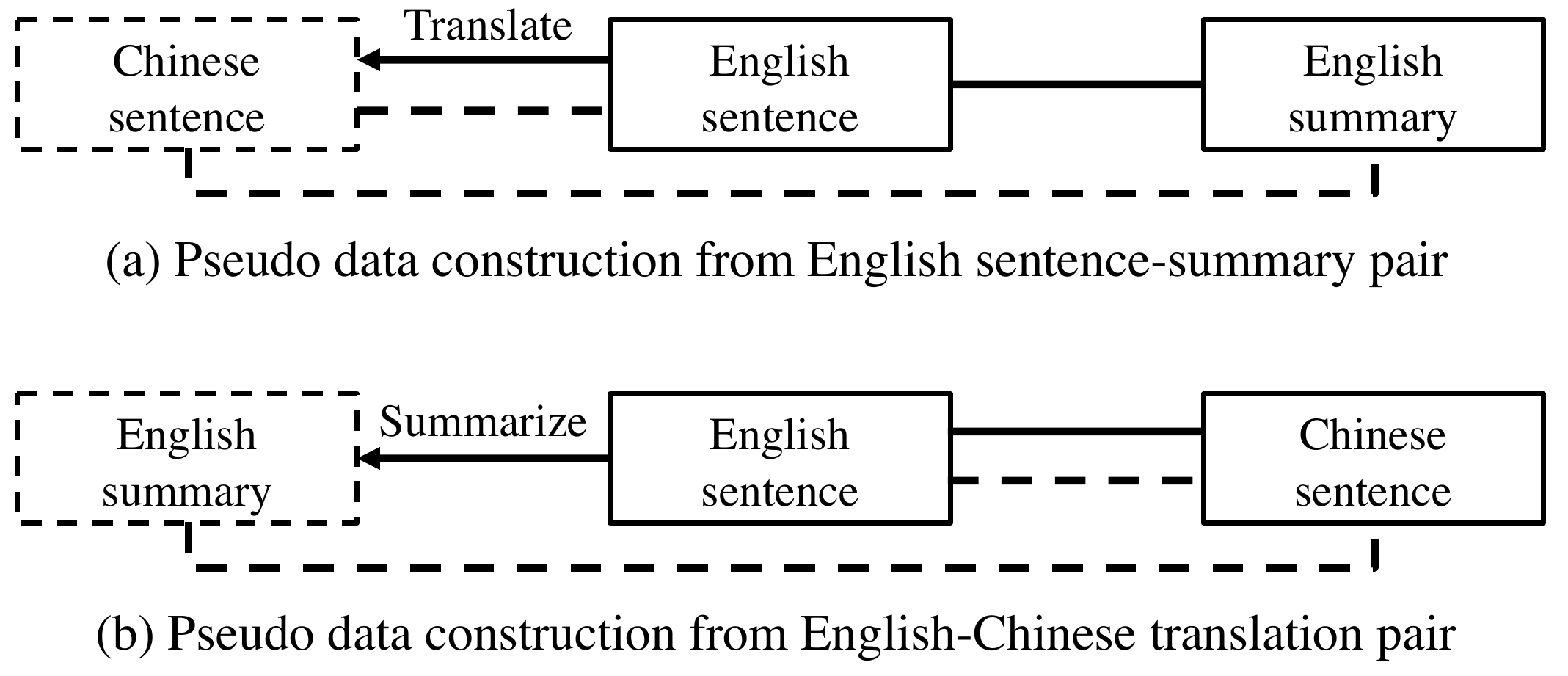}
   \caption{An example of pseudo data construction (Chinese-English abstractive summarization in this case) from parallel corpora. The solid line box denotes genuine data and the dashed line box denotes automatically generated pseudo data. The arrow indicates generation by the neural encoder-decoder model. The solid line denotes genuine parallel data and the dashed line denotes pseudo parallel data. In (a), we use the generated Chinese sentence for pseudo translation data in addition to pseudo Chinese-English summarization pairs. In (b), we also treat genuine translation pairs as cross-lingual summarization data without any compression.}
   \label{fig:data_construction}
\end{figure}

\newcite{duan-etal-2019-zero} constructed pseudo cross-lingual abstractive summarization data from monolingual abstractive summarization data by back-translation~\cite{sennrich-etal-2016-improving}.
In this study, we also construct pseudo training data.
In addition to monolingual summarization data, we use translation pairs as a source parallel corpus for pseudo data construction.
Figure \ref{fig:data_construction} shows an example of pseudo data construction for Chinese-English abstractive summarization.

\subsection{From Monolingual Summarization}
Figure \ref{fig:data_construction} (a) illustrates the process of constructing pseudo Chinese-English summarization data from English sentence-summary pairs.
Following \newcite{duan-etal-2019-zero}, we apply a neural machine translation model (English-to-Chinese in this figure) to the source sentences in the monolingual abstractive summarization dataset.
Then, we use the pairs of translated sentences and genuine summaries as pseudo cross-lingual sentence-summary pairs.

In addition, we use the pairs of translated sentences and genuine source sentences as pseudo translation pairs.
For these pairs, we attach a special token \texttt{<PseudoTrans>} to the source side sentence to indicate pseudo translation data such as \newcite{caswell-etal-2019-tagged}.

\subsection{From Translation Pairs}
Figure \ref{fig:data_construction} (b) illustrates the process of constructing pseudo Chinese-English summarization data from English-Chinese translation pairs.
As shown in this figure, we apply a monolingual abstractive summarization model to sentences in the target language (English in this figure) to generate summaries.
Then, we use the pairs of generated summaries and sentences in the source language as pseudo cross-lingual sentence-summary pairs.

Moreover, we can consider the genuine translation pairs as the cross-lingual sentence-summary pairs without any compression.
Thus, we also use the translation pairs as the cross-lingual sentence-summary pairs.

\section{Experiments}
We conduct experiments on three language pairs: Chinese-English, Arabic-English, and English-Japanese to evaluate the effectiveness of our proposed multi-task learning framework, Transum.
As described in Section \ref{sec:transum}, we used Transformer~\cite{NIPS2017_7181} as the neural encoder-decoder model and LRPE~\cite{takase-okazaki-2019-positional} to control output sequence length\footnote{\href{https://github.com/takase/control-length}{https://github.com/takase/control-length}}.
We constructed vocabulary based on the unigram language model~\cite{kudo-2018-subword}\footnote{\href{https://github.com/google/sentencepiece}{https://github.com/google/sentencepiece}}.
We set vocabulary size 32K for each language in each corpus.

\subsection{Baselines}
We compared the following methods with Trunsum to investigate the effect of our multi-task learning framework.

\paragraph{Translation}
We trained Transformer~\cite{NIPS2017_7181} on genuine translation pairs and then used it to translate the given sentence.
In other words, this method outputs translated sentences without any compression.

\paragraph{Translation+Summarization}
We trained Transformer on the monolingual abstractive summarization dataset.
Then we used it to summarize the sentence translated by the above Translation baseline.
Thus, this configuration represents the pipeline approach with the neural encoder-decoder.

\paragraph{Zero-shot}
As described in Section \ref{sec:transum}, our introduced special tokens \texttt{<Trans>} and \texttt{<Summary>} enable the neural encoder-decoder to process cross-lingual abstractive summarization without the corresponding parallel corpus.
Thus, we used only translation pairs and a monolingual abstractive summarization dataset for training.

\paragraph{Pseudo only}
In contrast to the zero-shot configuration, we trained Transformer with the pseudo cross-lingual abstractive summarization data only.
We constructed the pseudo data by applying the Translation and Summarization baselines to genuine parallel corpora.

\paragraph{Methods of recent studies}
We trained the methods proposed in recent studies for cross-lingual abstractive summarization~\cite{zhu-etal-2019-ncls,zhu-etal-2020-attend}\footnote{\href{https://github.com/ZNLP/NCLS-Corpora}{https://github.com/ZNLP/NCLS-Corpora}}\footnote{\href{https://github.com/ZNLP/ATSum}{https://github.com/ZNLP/ATSum}} on our used training data.
In particular, comparison with the method of \newcite{zhu-etal-2019-ncls} is important in evaluating our proposed method since their method also uses both of genuine and pseudo data for training.
In addition, we describe the reported scores in previous studies for Chinese-English and Arabic-English summarization.

\subsection{Dataset}\label{sec:dataset}
We describe genuine parallel corpora which are used for training each method and pseudo data construction.
Moreover, we describe test data for each language pair.

\paragraph{Chinese-English}
We used 0.9M Chinese-English sentence pairs extracted from LDC corpora\footnote{The corpora include LDC2004T07, Hansards portion of LDC2004T08, LDC2005T06, and LDC2015T06. We tried to obtain LDC2003E14 which is used in previous studies~\cite{wang-etal-2017-deep-neural,zhang-etal-2019-bridging} but it was unavailable because LDC has stopped its distribution. Thus, the number of translation pairs used in this study is less than the number used in previous studies.} as genuine translation pairs.
For English abstractive summarization, we used 3.8M sentence-summary pairs extracted from Annotated English Gigaword~\cite{napoles:2012:AG}\footnote{\href{https://catalog.ldc.upenn.edu/LDC2012T21}{https://catalog.ldc.upenn.edu/LDC2012T21}} by the pre-processing script of \newcite{rush-etal-2015-neural}\footnote{\href{https://github.com/facebookarchive/NAMAS}{https://github.com/facebookarchive/NAMAS}}.
The original script converts digits and infrequent words into the special token but we ignored this procedure in the same as the recent study~\cite{takase-okazaki-2019-positional}.

For evaluation, we used the Chinese-English summarization dataset constructed by \newcite{duan-etal-2019-zero}.
They manually translated the English source sentences in DUC 2004 task 1 data~\cite{Over:2007:DC:1284916.1285157} into Chinese.
The evaluation set contains 500 Chinese source sentences and four kinds of English reference summaries for each Chinese sentence.

\paragraph{Arabic-English}
We used 1.1M Arabic-English sentence pairs extracted from LDC corpora\footnote{The corpora include LDC2004T17, LDC2004T18, LDC2005T05, and LDC2007T08.} as genuine translation pairs.
For English abstractive summarization, we used the same dataset as in the Chinese-English experiment.

To evaluate the performance, we used the DUC 2004 task 3 dataset~\cite{Over:2007:DC:1284916.1285157}.
The dataset contains 240 English source sentences translated from Arabic documents and four kinds of English reference summaries for each source sentence.
We obtained the original Arabic sentences from the Agence France Presse portion of Arabic Gigaword\footnote{\href{https://catalog.ldc.upenn.edu/LDC2011T11}{https://catalog.ldc.upenn.edu/LDC2011T11}}, and then used the pairs of Arabic sentences and English summaries as the evaluation set.

\paragraph{English-Japanese}
In the above two configurations, we used English as the target language whereas we are interested in the case where English is the source language.
However, we could not find a cross-lingual abstractive summarization evaluation set containing English sentences as the source sentences with length constraints for each output summary.
To construct such an evaluation set, we manually translated the source sentences in a Japanese abstractive summarization dataset~\cite{Hitomi2019}, which contains source sentences and three kinds of summaries depending on length constraints, into English.
The constructed evaluation set contains 1,489 English sentences and three kinds of Japanese summaries for each English sentence.

For genuine English-Japanese translation pairs, we used the JIJI corpus\footnote{\href{http://lotus.kuee.kyoto-u.ac.jp/WAT/jiji-corpus/}{http://lotus.kuee.kyoto-u.ac.jp/WAT/jiji-corpus/}} which contains 0.2M sentence pairs.
Since this corpus is too small to train a neural encoder-decoder, we used JparaCrawl~\cite{morishita19jparacrawl}\footnote{\href{http://www.kecl.ntt.co.jp/icl/lirg/jparacrawl/}{http://www.kecl.ntt.co.jp/icl/lirg/jparacrawl/}} to augment the training data for the Japanese-English translation model.
For Japanese abstractive summarization, we used the Japanese News Corpus~\cite{Hitomi2019}\footnote{\href{https://cl.asahi.com/api_data/jnc-jamul-en.html}{https://cl.asahi.com/api\_data/jnc-jamul-en.html}}, which contains 1.9M sentence-summary pairs.

\subsection{Results of Cross-Lingual Abstractive Summarization}
\begin{table}[!t]
  \centering
  \scriptsize
  \begin{tabular}{| l | c c c |} \hline
  Method                        & R-1 & R-2 & R-L \\ \hline
  \newcite{8370729}             & 19.3 \ \ & 4.3 \ \ & 17.0 \ \ \\
  \newcite{ijcai2017-555}       & 24.0 \ \ & 7.0 \ \ & 21.3 \ \ \\
  \newcite{duan-etal-2019-zero} & 26.0 \ \ & 8.0 \ \ & 23.1 \ \ \\ \hline
  Translation                   & 14.96 & 3.01 & 13.17 \\
  + Summarization               & 18.82 & 5.08 & 17.00 \\
  Zero-shot                & 19.20 & 4.92 & 17.01 \\ 
  Pseudo only              & 24.72 & 7.37 & 21.29 \\
  \newcite{zhu-etal-2019-ncls}  & 19.92 & 6.08 & 17.91 \\ 
  \newcite{zhu-etal-2020-attend} & 18.74 & 5.46 & 17.03 \\
  Transum (proposed)             & {\bf 26.69} & {\bf 8.49} & {\bf 23.33} \\ \hline
  \end{tabular}
  \caption{Recall-oriented ROUGE scores of each method in Chinese-English abstractive summarization. The upper part of this table shows scores reported in previous studies\protect\footnotemark.}
  \label{tab:zhen_headline}
\end{table}
\footnotetext{We exclude the scores of \citet{DBLP:conf/aaai/CaoWLL18} because they did not report recall-oriented ROUGE scores but F-1 ROUGE scores.}

\begin{table}[!t]
  \centering
  \scriptsize
  \begin{tabular}{| l | c c c |} \hline
  Method                        & R-1 & R-2 & R-L \\ \hline
  \newcite{Doran04newsstory}    & 25.87& 4.73 & 21.98 \\
  \newcite{ouyang-etal-2019-robust}& 29.43 & 7.02 & 19.89 \\ \hline
  Translation                   & 29.57 & 10.01 & 26.21 \\
  + Summarization               & 31.85 & 10.98 & 28.48 \\
  Zero-shot                     & 29.89 & 10.34 & 26.87 \\
  Pseudo only                   & 35.78 & 11.98 & 31.30 \\
  \newcite{zhu-etal-2019-ncls}    & 31.38 & 11.29 & 28.35 \\
  \newcite{zhu-etal-2020-attend}  & 29.43 & 10.33 & 26.75 \\
  Transum (proposed)              & {\bf 36.04} & {\bf 12.12} & {\bf 31.49} \\ \hline
  \end{tabular}
  \caption{Recall-oriented ROUGE scores of each method in Arabic-English abstractive summarization. The upper part of this table shows scores reported in previous studies.}
  \label{tab:aren_headline}
\end{table}

\begin{table*}[!t]
  \centering
  \scriptsize
  \begin{tabular}{| l | c c c | c c c | c c c |} \hline
    & \multicolumn{3}{|c|}{$L = 10$}&\multicolumn{3}{c|}{$L = 13$}&\multicolumn{3}{c|}{$L = 26$} \\ \hline
  Method                        & R-1 & R-2 & R-L & R-1 & R-2 & R-L & R-1 & R-2 & R-L \\ \hline
  Translation                 & 9.31 & 1.41 & 8.89 & 13.16 & 1.91 & 11.87 & 18.88 & 2.81 & 14.51 \\
  + Summarization             & 15.25 & 3.89& 14.63 & 19.64& 5.84 & 18.28 & 21.72 & 5.25 & 17.65 \\
  Zero-shot                   & 11.40 & 2.82 & 11.14 & 14.76 & 3.61 & 13.77 & 24.61 & 5.84 & 19.17 \\
  Pseudo only                 & 22.35 & 7.68 & 21.73 & 26.73 & 9.43 & 25.19 & 32.65 & 10.28& 25.89  \\
  \newcite{zhu-etal-2019-ncls}  & 17.82 & 6.01 & 15.88 & 22.38 & 7.66 & 19.41 & 27.71 & 9.85 & 23.23 \\
  \newcite{zhu-etal-2020-attend} & 16.71 & 5.22 & 14.93 & 20.41 & 6.72 & 17.89 & 23.85 & 8.13 & 20.41 \\
  Transum (proposed) & {\bf 27.17} & {\bf 10.92} & {\bf 26.44} & {\bf 32.87} & {\bf 13.75} & {\bf 30.87} & {\bf 38.83} & {\bf 14.45} & {\bf 30.99} \\ \hline
  \end{tabular}
  \caption{Recall-oriented ROUGE scores of each method in English-Japanese abstractive summarization. The English-Japanese test set contains three kinds of Japanese summaries depending on the desired length $L = 10$, $13$, and $26$. Thus, we report the scores for each desired length.}
  \label{tab:enja_headline}
\end{table*}

\begin{table}[!t]
  \centering
  \scriptsize
  \begin{tabular}{| l | c c c |} \hline
  Method                        & R-1 & R-2 & R-L \\ \hline
  \newcite{8370729}             & 21.5 \ \ & 6.6 \ \ & 19.6 \ \ \\
  \newcite{ijcai2017-555}       & 26.7 \ \ & 10.2 \ \ & 24.3 \ \ \\
  \newcite{duan-etal-2019-zero} & 30.1 \ \ & 12.2 \ \ & 27.7 \ \ \\
  \newcite{DBLP:conf/aaai/CaoWLL18} & {\bf 32.04} & 13.60 & 27.91 \\ \hline
  \newcite{zhu-etal-2019-ncls}    & 29.61 & 12.55 & 27.41 \\
  \newcite{zhu-etal-2020-attend}  & 28.19 & 11.89 & 26.28 \\
  Transum w/o LRPE              & 31.43 & {\bf 13.99} & {\bf 29.04} \\ \hline
  \end{tabular}
  \caption{The full-length F-1 based ROUGE scores of each method in the Chinese-English abstractive summarization dataset constructed from English Gigaword. The upper part of this table shows scores reported in previous studies.}
  \label{tab:zhen_gigaword}
\end{table}

\begin{table*}[!t]
  \centering
  \scriptsize
  \begin{tabular}{| c c | c c c | c || c | c | c |} \hline
  \multicolumn{6}{|c||}{Training data} &  MT02 & \multicolumn{2}{|c|}{DUC} \\ \hline
  \multicolumn{2}{|c|}{Genuine} & \multicolumn{3}{c|}{Pseudo (cross-lingual sum)} & Pseudo & Zh-En & En-En & Zh-En \\ 
  Trans & Sum & From trans & From sum & Trans as sum & (trans) & BLEU & R-1 & R-1 \\ \hline
  \multicolumn{9}{|c|}{Cross-lingual abstractive summarization with zero-shot} \\ \hline
 \Checkmark &          &         &                &     &         & 40.86& - & 14.96 \\
 \Checkmark & \Checkmark &         &                &     &         &41.87&30.36&17.77 \\
 \Checkmark & \Checkmark &        &           &\Checkmark&          &43.80&30.16&17.40 \\
 \Checkmark & \Checkmark &        &           &          &\Checkmark&47.87&29.97&19.20\\
 \Checkmark & \Checkmark &        &           &\Checkmark&\Checkmark&{\bf 49.31}&30.11&18.06\\ \hline
 \multicolumn{9}{|c|}{Trained with pseudo cross-lingual abstractive summarization dataset} \\ \hline
 \Checkmark & \Checkmark &\Checkmark&        &             &         &43.99&29.16&24.68 \\
 \Checkmark & \Checkmark &\Checkmark&\Checkmark      &     &         &44.48&29.82&25.42 \\
 \Checkmark & \Checkmark &\Checkmark&\Checkmark&\Checkmark&         &45.26&29.68&26.05 \\
 \Checkmark & \Checkmark &\Checkmark&\Checkmark&\Checkmark &\Checkmark&48.64&{\bf 30.37}&{\bf 26.69}\\ \hline
  \end{tabular}
  \caption{Results of the model trained with each training dataset. This table shows BLEU scores in Chinese-English (Zh-En) MT02 and ROUGE-1 scores in the original DUC 2004 task 1 and Chinese-English DUC 2004. \label{tab:ablation}}
\end{table*}

Tables \ref{tab:zhen_headline}, \ref{tab:aren_headline}, and \ref{tab:enja_headline} show the recall-oriented ROUGE scores of each method in the Chinese-English, Arabic-English, and English-Japanese abstractive summarization test sets respectively.
For Chinese-English and Arabic-English, we truncated characters over 75 bytes in generated summaries before computing ROUGE scores following the DUC 2004 evaluation protocol~\cite{Over:2007:DC:1284916.1285157}.
For English-Japanese, we truncated characters exceeding each desired length $L$ before ROUGE computation.
These tables contain ROUGE-1, 2, and L scores (R-1, R-2, and R-L respectively).

These tables indicate that the proposed Transum achieved the top score in each test set.
These results show that our proposed multi-task learning framework had a positive influence on the cross-lingual abstractive summarization task.
In particular, Transum outperformed Pseudo only approach.
This result indicates that it is effective to use genuine translation pairs and monolingual summarization datasets.

The pseudo only approach achieved better performance than the zero-shot method in all language pairs.
This result indicates that we should train the encoder-decoder with the dataset corresponding to the target task even if the training dataset is automatically constructed.

The zero-shot approach achieved comparable ROUGE scores to Translation except for Chinese-English and English-Japanese $L=26$.
This result implies that it is difficult to process the task in the zero-shot configuration.
In contrast, the pipeline approach, which is a combination of machine translation and summarization, outperformed the zero-shot method in most test sets.
Thus, it is better to combine the models trained for specific tasks than to address the cross-lingual abstractive summarization in zero-shot.

Tables \ref{tab:zhen_headline}, \ref{tab:aren_headline}, and \ref{tab:enja_headline} show that Transum outperformed the recent methods~\cite{zhu-etal-2019-ncls,zhu-etal-2020-attend}.
However, these methods cannot control output sequence length and thus these comparisons might be unfavorable for them.
Therefore, we compared these methods with Transum without the length control method (LRPE) in an additional test set.
We used the Chinese-English summarization test set constructed from Annotated English Gigaword~\cite{napoles:2012:AG} by \newcite{duan-etal-2019-zero}.
Table \ref{tab:zhen_gigaword} shows F-1 based ROUGE scores of each method.
This table indicates that Transum achieved better scores than those of \newcite{zhu-etal-2019-ncls} and \newcite{zhu-etal-2020-attend}.
These results show that our multi-task learning framework offers a better approach than methods in these studies.

\begin{figure}[!t]
  \centering
  \includegraphics[width=6.5cm]{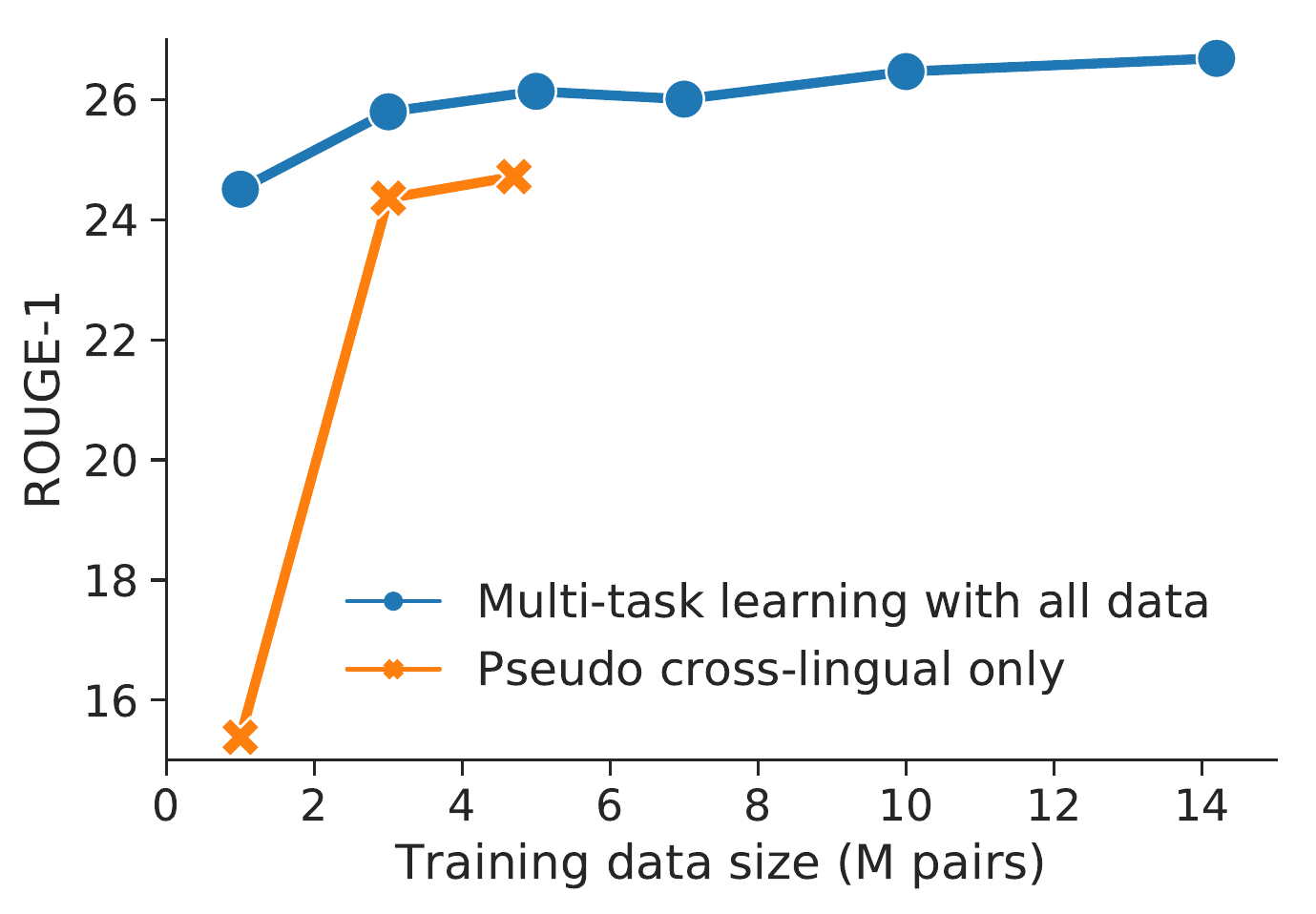}
   \caption{ROUGE-1 scores in the Chinese-English abstractive summarization test set. This figure reports the scores of models trained with varying the number of training pairs.}
   \label{fig:score_by_datasize}
\end{figure}

We investigate details of the contribution of each training data to the performance.
Table \ref{tab:ablation} shows the performance in Chinese-English summarization in the case where we varied the kinds of datasets used to train the encoder-decoder.
This table shows BLEU scores in the NIST 2002 Chinese-English translation test set and ROUGE-1 scores in the DUC 2004 task 1 (English abstractive summarization test set) in addition to ROUGE-1 scores in Chinese-English summarization.
This table is composed of two parts: without and with constructed pseudo cross-lingual abstractive summarization data for training.

Table \ref{tab:ablation} indicates that the more training data we use, the better performance we achieve in Chinese-English translation and summarization.
The upper part of this table shows that we improved ROUGE scores in Chinese-English abstractive summarization even though we did not use training data corresponding to this task.
In particular, addition of pseudo translation pairs (4th row) raised more than four ROUGE-1 score compared to the case using only genuine translation pairs.
These results imply that training with (pseudo) translation pairs has a positive effect on cross-lingual abstractive summarization.
Moreover, we improved BLEU score more than eight points.

\begin{table}[!t]
  \centering
  \scriptsize
  \begin{tabular}{| l | c c c c |} \hline
  Method                               &MT02  & MT03  & MT04  & MT05  \\ \hline
  \newcite{wang-etal-2017-deep-neural} & -     & 39.35 & 41.15 & 38.07 \\ 
  \newcite{cheng-etal-2018-towards}    & 46.10 & 44.07 & 45.61 & 44.06 \\ 
  \newcite{cheng-etal-2019-robust}     & 48.13 & 47.83 & 49.13 & {\bf 49.04}\\ 
  \newcite{zhang-etal-2019-bridging}   & -     & 48.31 &{\bf 49.40} & 48.72 \\ \hline 
  Transformer                          & 40.86    & 40.71     & 40.51 & 37.22 \\
  Transum (proposed)                  &{\bf 48.64}&{\bf 50.23}& 48.45 & 47.84 \\ \hline
  \end{tabular}
  \caption{BLEU scores of each method in the NIST Chinese-English test set. \label{tab:mt_zhen}}
\end{table}

\begin{table}[!t]
  \centering
  \scriptsize
  \begin{tabular}{| l | c c c c |} \hline
  Method                               &MT02  & MT03  & MT04  & MT05  \\ \hline 
  \newcite{DBLP:journals/corr/AlmahairiCHC16} & - & - & - & 51.19 \\ 
  \newcite{DBLP:journals/corr/abs-1808-06116} & - & - & - & 58.35 \\ \hline
  Transformer                          &50.54 & {\bf 62.37} & 50.89 & 52.96 \\
  Transum (proposed)                   &{\bf 53.34} & 60.94 & {\bf 55.53} & {\bf 58.57} \\ \hline
  \end{tabular}
  \caption{BLEU scores of each method in the NIST Arabic-English test set. \label{tab:mt_aren}}
\end{table}

\begin{table}[!t]
  \centering
  \footnotesize
  \begin{tabular}{| l | c |} \hline
  Method                               & BLEU  \\ \hline
  \newcite{morishita-etal-2017-ntt}    & 19.41 \\
  \newcite{susanto-etal-2019-sarahs}   & 21.91 \\ \hline
  Transformer                          & 17.62 \\
  Transum (proposed)                   & {\bf 23.10}\\ \hline
  \end{tabular}
  \caption{BLEU scores of each method in the JIJI English-Japanese test set. \label{tab:mt_enja}}
\end{table}

The lower part of Table \ref{tab:ablation} shows that the pseudo cross-lingual abstractive summarization dataset significantly improves ROUGE-1 scores in Chinese-English summarization.
These results indicate that we need the dataset corresponding to the target task to achieve better scores.

In contrast to other tasks, we cannot find significant improvement for ROUGE-1 scores in the monolingual abstractive summarization task.
In other words, our multi-task learning framework had little effect on monolingual abstractive summarization.
We consider the reason is that we did not prepare an additional monolingual dataset for training.
Thus, we might improve its performance by adding any monolingual task such as paraphrasing.

Figure \ref{fig:score_by_datasize} illustrates ROUGE-1 scores in the Chinese-English summarization dataset when we varied the training data size.
This figure contains two configurations: using the pseudo cross-lingual abstractive summarization data only and using both genuine and pseudo data.
This figure shows that we can achieve better performance in using both genuine and pseudo data to train the model.
This result indicates that multi-task learning is a better approach than using the training data corresponding to the target task only.

The ROUGE-1 score of the model trained with pseudo cross-lingual only is low when the training data size is small.
Since this dataset contains only summarization, the number of tokens for generation is smaller than that in translation pairs.
Thus, it might be difficult to learn the alignment of tokens in a source sentence with output tokens.

\subsection{Results of Machine Translation}
Since Table \ref{tab:ablation} shows that our Transum also improved the BLEU score in Chinese-English translation, we further investigate the performance of Transum in machine translation.
For Chinese-English and Arabic-English, we used NIST 2002, 2003, 2004, and 2005 test sets.
We computed BLEU scores with the official tool.
Moreover, we used the test set of the JIJI corpus for English-Japanese.

Tables \ref{tab:mt_zhen}, \ref{tab:mt_aren}, and \ref{tab:mt_enja} show BLEU scores of Transformer trained with only genuine translation pairs (i.e., the Translation baseline) and our Transum in Chinese-English, Arabic-English and English-Japanese respectively.
Each table also indicates BLEU scores reported in previous studies\footnote{Each table includes scores of single models. For \newcite{DBLP:journals/corr/abs-1808-06116}, we focused on the setting which uses almost the same translation pairs as ours.}.

These tables show that Transum outperformed Transformer trained with only translation pairs except for MT03 in Arabic-English.
Thus, our multi-task learning approach also has a positive effect on machine translation.
In particular, Transum raised the BLEU score approximately eight or more points compared to the Transformer trained with only translation pairs in all test sets of Chinese-English translation.
Moreover, Transum outperformed the previous top score on MT02 and MT03 in Chinese-English and JIJI English-Japanese translation even though Transum has no specific approach to machine translation.
Furthermore, since the methods proposed in \newcite{zhang-etal-2019-bridging} and \newcite{cheng-etal-2019-robust} are orthogonal to Transum, we expect further improvement in machine translation by introducing their methods into Transum.

\section{Related Work}
Early explorations on cross-lingual summarization adopted the pipeline approach, which combines the machine translation with summarization methods~\cite{leuski2003cross-lingual,orasan-chiorean-2008-evaluation,wan-etal-2010-cross,yao-etal-2015-phrase}.
\newcite{orasan-chiorean-2008-evaluation} applied the Maximal Marginal Relevance to summarize given documents and then automatically translated the summary.
To prevent unreadable outputs, \newcite{wan-etal-2010-cross} proposed the method to predict the machine translation quality for sentences in a source document, and then generate a summary based on the predicted quality score before translation.
\newcite{yao-etal-2015-phrase} extended phrase-based machine translation models to select important phrases.
However, \newcite{wan-2011-using} indicated that we should use information from both sides rather than such pipeline approaches.

Recent studies applied a neural encoder-decoder model to generate cross-lingual abstractive summaries from the given document directly~\cite{8370729,duan-etal-2019-zero,zhu-etal-2019-ncls}.
To construct pseudo cross-lingual abstractive summarization data for training, \newcite{8370729} adopted the approach consisting of two steps: machine translation of a document in the source language and then monolingual abstractive summarization of the translated document.
Then, they used the pairs of summarized translations and the original documents as pseudo training data.
\newcite{duan-etal-2019-zero} used genuine summaries to improve the quality of constructed training data.
They applied a machine translation model to source sentences in monolingual sentence-summary pairs and used the pairs of translated sentences and genuine summaries as pseudo training data.
\newcite{zhu-etal-2019-ncls} proposed a round-trip translation strategy to obtain high quality pseudo training data from existing monolingual summarization datasets.
Their round-trip strategy translates a source sentence in monolingual sentence-summary pairs in the same manner as \newcite{duan-etal-2019-zero}, and then re-translates the translated sentence into the source language.
Their approach filters out based on the similarity between the source sentence and round-trip translation.
These studies explored the sophisticated way to construct pseudo training data but pay little attention to the existing genuine parallel corpora.
In contrast, this study utilizes such genuine parallel corpora in addition to constructed pseudo data with the multi-task learning framework.

In addition to the round-trip translation strategy, \newcite{zhu-etal-2019-ncls} introduced a multi-task learning approach for cross-lingual abstractive summarization.
Their method prepares two decoders: one for cross-lingual abstractive summarization and the other for machine translation or monolingual summarization.
The method trains the decoders to generate corresponding output for a given sentence.
They indicated that their multi-task learning approach improved the performance of cross-lingual abstractive summarization but it requires additional parameters for a decoder.
In contrast, our proposed Transum is more simple because it only needs to attach the special token to the source sentence.
Moreover, experimental results show that Transum outperformed the multi-task approach of \newcite{zhu-etal-2019-ncls}.

\section{Conclusion}
This paper presents a multi-task learning framework for cross-lingual abstractive summarization to augment training data.
The proposed method, Transum, attaches the special token to the beginning of the input sentence to indicate the target task.
The special token enables us to use genuine translation pairs and the monolingual abstractive summarization dataset in addition to the pseudo cross-lingual abstractive summarization data for training.
The experimental results show that Transum achieved better performance than the pipeline approach and model trained with pseudo data only.
We achieved the top ROUGE scores in Chinese-English and Arabic-English abstractive summarization.
Moreover, Transum also improved the performance of machine translation and outperformed the previous top score in the JIJI English-Japanese translation.

\bibliographystyle{acl_natbib}
\bibliography{emnlp2020.bbl}

\newpage
\appendix
\section{Details of Experimental Settings}
To train Transum, we used the same hyperparameters as the described values in the original implementation\footnote{\href{https://github.com/takase/control-length}{https://github.com/takase/control-length}} except for the mini-batch size.
We varied the mini-batch size in proportion to the training data size.
For example, if the whole training data size is twice as much as English abstractive summarization training data size, we double the mini-batch size from the size in training English abstractive summarization.
In other words, we did not search appropriate hyperparameters for our Transum.
Moreover, the number of parameters in Transum is the identical to one of Transformer in the base setting~\cite{NIPS2017_7181} except for word embeddings.
We used Tesla P100 GPUs for our experiments.
In decoding summaries, we tuned the desired length corresponding to the length constraint in each test set.

\end{document}